\documentclass[preprint,10pt]{elsarticle}

\usepackage{amssymb}
\usepackage{amsmath}
\usepackage{orcidlink}
\usepackage{changepage}
\usepackage{float}

\usepackage[capitalize]{cleveref}
\crefname{figure}{Fig.}{Figs.}

\usepackage{geometry}
\usepackage{tikz}
\usepackage{booktabs} 

\usepackage{ragged2e}
\AtBeginDocument{\RaggedRight}

\biboptions{sort&compress}

\let\oldthebibliography\thebibliography

\renewcommand{\thebibliography}[1]{%
  \oldthebibliography{#1}%
  \setlength{\itemsep}{1pt}
  \setlength{\parskip}{0pt}
}
\usepackage{todonotes}
\newboolean{SHOWCOMMENTS}
\setboolean{SHOWCOMMENTS}{true}
\ifthenelse{\boolean{SHOWCOMMENTS}}{
\newcommand{\tdMK}[1]
{\todo[color=blue!25,inline]{\footnotesize{\bf Martin:} #1}}
}{
\newcommand{\tdMK}[1]{}
}
\newcommand{\MK}[1]{{\color{blue}#1}}

\begin{document}

\makeatletter
\def\ps@pprintTitle{}
\makeatother

\begin{frontmatter}



\title{Graph Neural Network Surrogates to leverage Mechanistic Expert Knowledge towards Reliable and Immediate Pandemic Response}


\author[1,first]{Agatha Schmidt\orcidlink{0009-0006-5766-8804}}
\fntext[first]{These authors contributed equally to this work.}
\author[1,first]{Henrik Zunker\orcidlink{0000-0002-9825-365X}}
\author[2]{Alexander Heinlein\orcidlink{0000-0003-1578-8104}}
\author[1,3]{Martin J. Kühn\orcidlink{0000-0002-0906-6984}\corref{cor1}}

\cortext[cor1]{Corresponding author. E-mail: martin.kuehn@dlr.de}

\affiliation[1]{organization={Institute of Software Technology, Department of High-Performance Computing, German Aerospace Center}, 
               city={Cologne}, 
               country={Germany}}

\affiliation[2]{organization={Delft Institute of Applied Mathematics, Delft University of Technology}, 
               city={Delft}, 
               country={The Netherlands}}

\affiliation[3]{organization={Life and Medical Sciences Institute and Bonn Center for Mathematical Life Sciences, University of Bonn}, 
               city={Bonn}, 
               country={Germany}}

%% Abstract
\begin{abstract}
During the COVID-19 crisis, mechanistic models have guided evidence-based decision making. However, time-critical decisions in a dynamical environment limit the time available to gather supporting evidence.
We address this bottleneck by developing a graph neural network (GNN) surrogate of an \MK{age-structured and} spatially resolved mechanistic metapopulation \MK{simulation model}. This combined approach \MK{complements classical modeling approaches which are mostly mechanistic and purely data-driven} machine learning approaches which are often black box. Our design of experiments spans outbreak and persistent-threat regimes, up to three contact change points, and age-structured contact matrices on a spatial graph \MK{with 400 nodes representing German counties}. We benchmark multiple GNN layers and identify an ARMAConv-based architecture that offers a strong accuracy–runtime trade-off. Across \MK{ horizons of 30--90 day simulation and prediction, allowing} up to three contact change points, the surrogate \MK{model} attains 10--27\% mean absolute percentage error (MAPE) while delivering (near) constant runtime with respect to the forecast horizon. Our approach accelerates evaluation by up to 28\,670 times compared with the mechanistic model, allowing responsive decision support in  time-critical scenarios and straightforward web integration. These results show how GNN surrogates can translate complex metapopulation models into immediate, reliable tools for pandemic response.
\end{abstract}

% \includegraphics[scale=0.3]{graphical_abstract/graphical_abstract.png}

% \begin{highlights}
% \item GNN surrogate matches metapopulation expert model at 10--27\% MAPE
% \item Up to 28\,670 times faster; less than $0.1\,s$ per 100 predictions
% \item Extensive grid search over 80 models, ARMAConv (7$\times$512) yields best accuracy
% \item Interventions modeled through three different contact change points
% \item Integration of mobility adjacency matrix
% \end{highlights}

%% Keywords

\end{frontmatter}

%\linenumbers

\section{Introduction}
\label{sec:intro}
Mathematical models can help public health experts and decision makers to explore potential future outcomes of ongoing disease dynamics. In particular during the recent COVID-19 crisis, mathematical modeling has been one of the principal forms to provide evidence on the effectiveness of public health and social interventions~\cite{ecdc_2024}. With an estimated number of more than 600\,000 undiscovered viruses in mammal and avian hosts that are capable of infecting humans~\cite{daszak_workshop_2020}, pandemic preparedness is a necessity.

Due to local outbreaks and human contact patterns, infectious disease dynamics are often heterogeneous on a spatial or demographic scale. For efficient mitigation of pandemics and accurate predictions, local transmission dynamics should be considered and corresponding data should be integrated in a mathematical model. 
Over the last years, a large number of authors have made contributions to predict the development of SARS-CoV\nobreakdash-2. Simple ordinary differential equation (ODE)-based SIR (Susceptible-Infected-Removed)-type models~\cite{bauer_relaxing_2021,schuler_data_2021} have been used for their efficiency in time-critical moments. For using more realistic, nonexponential disease stage transitions, \MK{linear chain trick (see, e.g.,~\cite{macdonald_time_1978,PLOTZKE2026823}), generalized linear chain trick~\cite{hurtado_generalizations_2019}, or flexible integro-differential (see, e.g.,~\cite{medlock_spreading_2003,WENDLER2026129636})} equation-based models are available.

Through metapopulation modeling, various authors addressed the spatially heterogeneous spread of SARS-CoV\nobreakdash-2 and modeled entire countries with high efficiency and small execution times~\cite{pei_differential_2020,chen_compliance_2021,levin_effects_2021,liu_modelling_2022,zunker_novel_2024, ZUNKER2025116782}. \MK{For instance, Pei et al.~\cite{pei_differential_2020} considered effects of intervention timing on COVID-19 spread in the United States (US), Chen et al.~\cite{chen_compliance_2021} considered compliance and containment of COVID-19 across towns in the US, Levin et al.~\cite{levin_effects_2021} considered effects of short-term travel on COVID-19 spread in Minnesota, while Liu et al.~\cite{liu_modelling_2022} considered COVID-19 cases in Singapore and Zunker et al.~\cite{zunker_novel_2024,ZUNKER2025116782} considered travel-time aware and behavior-driven metapopulation models with the 400 German counties.}
To include individual transmission and superspreading events in an intuitive way, different contact network-, agent-, or individual-based models have been developed; see, e.g.,~\cite{bershteyn_implementation_2018,kerr_covasim_2021,muller_predicting_2021,KERKMANN2025110269}. \MK{For instance, Bershteyn et al. provided a very general pre-COVID-19 agent-based model (ABM) with different transmission pathways (vector, environmental, or airborne), Kerr et al.~\cite{kerr_covasim_2021} developed a powerful ABM for COVID-19, Müller et al.~\cite{muller_predicting_2021} extended the MATSim ABM for the COVID-19 use case, and Kerkmann et al.~\cite{KERKMANN2025110269} developed an ABM for respiratory diseases and direct or airborne transmission pathways.}
However, while more complex models allow more detailed research questions to be answered, they generally come at higher computational costs. Recent works with very similar approaches~\cite{bicker_hybrid_2024,kehrer_hybrid_2025,bostanci2024} combine agent- and equation-based metapopulation approaches to substantially reduce the prediction time of complex \MK{ABMs}, yet these models remain the most computationally demanding among the involved approaches.
Furthermore, many models using artificial intelligence (AI) have been proposed to fight the SARS-CoV\nobreakdash-2 pandemic\MK{; see, e.g.,~\cite{tayarani_n_applications_2021}}. While these models often deliver accurate results in short execution times, a review on more than 650 AI-based models identified most of them to be black box models~\cite{tayarani_n_applications_2021}. Unfortunately, black box models often lack interpretability or explainability and do not contribute to a better understanding of underlying processes—important when deciding and communicating interventions to the public.
Nevertheless, time is a critical factor in pandemics, and evaluations of mitigation and reaction strategies have to be conducted in narrow time windows. In addition, interventions ranging from distancing recommendations to face masks (in particular locations), capacity-restrictions, and closures of venues or even vaccinations (if available) yield a tremendously large combinatorial space from which decision makers need to select. While very simplified models are executed within the range of (milli)seconds, complex models, even if highly optimized, might need seconds to minutes to be run. With automated pipelines preventing manual delay and high-performance computing~\cite{memon_automated_2024}, thousands or millions of outcomes can be computed on a supercomputing facility to inform decision makers \MK{on short term}. However, in a dynamically changing environment, decision makers might need expected outcomes for \MK{individual} combinations of nonpharmaceutical interventions (NPIs) \MK{on-the-fly}. \MK{Training a GNN on the HPC simulations eventually allows the user-driven exploration of any strictness of NPIs, based on the interpolation capabilities of the GNN. Through a low-barrier access for public health experts, e.g., a web application, this approach also drastically increases the impact of the developed (mechanistic) models. Furthermore, novel NPIs, public recommendations, or observed behavioral changes will be represented in the mechanistic models with new contact patterns and learned regularly with new simulation outputs by novel versions of the GNN. }

\begin{figure}
\begin{adjustwidth}{-.25in}{-.25in}
\centering
\includegraphics[scale=0.25,trim={2.5cm 0 0 0},clip]{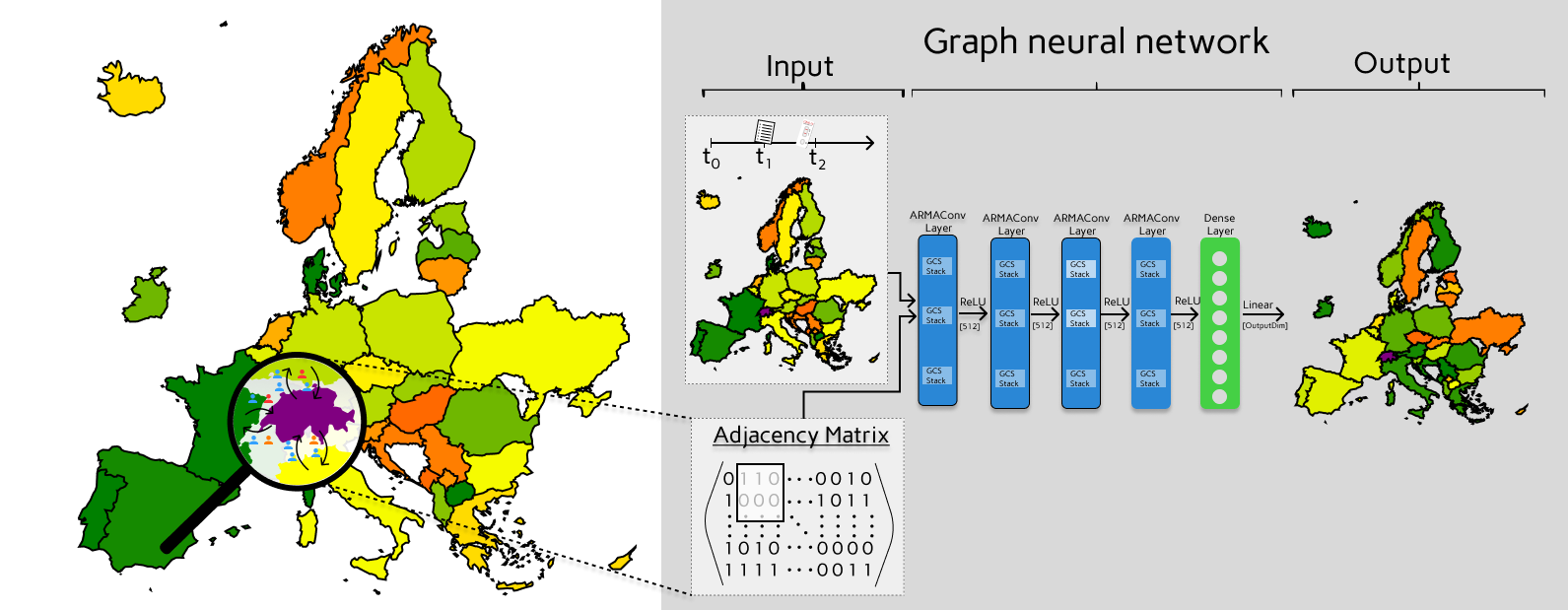}
\caption{\textbf{Visual representation of the prediction phase for a GNN surrogate model.} The adjacency matrix results from the links between the considered regions, here depicted high-level as European countries.  The magnifying glass highlights mobility connections between Switzerland and other countries. Colors on the national level geography represent disease outcomes such as symptomatic cases incidence, hospitalization, or deaths. Visualization of a GNN with four ARMAConv layers with three Graph Convolutional Skip (GCS) stacks and one dense output layer.
Visualization based on Eurostat shape files~\cite{eurostat} \textcopyright~EuroGeographics for the administrative boundaries.
}
\label{fig:graphical_abstract}
\end{adjustwidth}
\end{figure}

In this paper, we combine an already validated \MK{age-structured and} spatially resolved expert mechanistic model\MK{~\cite{kuhn_assessment_2021,koslow_appropriate_2022}} with AI-based approaches to provide an on-the-fly execution of spatially resolved infectious disease models. A schematic presentation of the approach is shown in~\cref{fig:graphical_abstract}. In~\cite{angione_using_2022}, \MK{Angione et al.} compared advantages and disadvantages of several established machine learning approaches such as random forests, decision trees, and neural networks as surrogates for agent-based models. Recently, \MK{Robertson et al.}~\cite{robertson_bayesian_2024} employed random forests to calibrate a city-scaled ABM. Here, however, we build upon an already validated \MK{age-structured and} spatially resolved metapopulation model~\cite{kuhn_assessment_2021}, whose simulations are used to train a machine learning model with spatial resolution.

The model proposed by \MK{Kühn et al.}~\cite{kuhn_assessment_2021} interprets the metapopulation model as a graph, with nodes representing subpopulations and edges capturing the exchange between them. With the expert model executed millions of times on a supercomputing facility overnight, we obtain a diverse training set from which a neural network surrogate can learn and generalize. As the output of the model from~\cite{kuhn_assessment_2021} is already in the form of a graph, a graph-based machine learning model is a natural choice and we therefore propose to employ graph neural networks (GNNs). GNNs are powerful tools for spatially distributed data, which have been introduced in \MK{Gori et al.}~\cite{gori_new_2005} and further popularized in \MK{Scarselli et al.}~\cite{scarselli_graph_2009}. While we propose GNNs as surrogates for expert mechanistic models of infectious disease dynamics, GNNs have already excelled on data in different applications such as weather predictions~\cite{Lam2023}, influenza-like diseases~\cite{deng_graph_2019}, or even SARS-CoV\nobreakdash-2~\cite{kapoor_examining_2020,gao_stan_2021,panagopoulos_transfer_2021,Fritz2022}. \MK{Fritz et al.~\cite{Fritz2022} used GNNs in a purely data-driven fashion for spatio‐temporal disease models, Gao et al.~\cite{gao_stan_2021} used patients' claims data from different counties in a graph-based approach with a transmission dynamics loss, and Panagopoulos et al.~\cite{panagopoulos_transfer_2021} used transfer learning together with reported case data-driven GNNs.}

In summary, we present a structured design of experiments covering outbreak and persistent‑threat regimes using age‑structured contacts and multiple contact‑change points. We introduce a compact spatio‑temporal input encoding for GNN surrogates and benchmark layers and architectures to identify an ARMAConv model with a strong accuracy–runtime trade‑off that, eventually, demonstrates large speed‑ups which enable interactive scenario exploration.

\section{Material and methods}
\label{sec:methods}

In this section, we detail the mechanistic simulator that provides the training targets, introduce the evaluation metric, describe the generation of input data, and document the neural network architectures used in the surrogate models. 

\subsection{Mechanistic model} 

\MK{A requirement for obtaining a surrogate model that is reliable in practice is that the model, which is replaced by the surrogate, has been validated beforehand. For infectious disease dynamics, t}he underlying mechanistic expert model that generates the reference training has to integrate the most important disease states and properties of disease dynamics for the surrogate model to learn from.   
While deaths are usually monitored, several indicators such as the number of reported cases, the number of hospitalizations or intensive care unit (ICU) admissions have been considered in Germany to implement or lift nonpharmaceutical interventions against SARS-CoV\nobreakdash-2~\cite{rki_control_covid_2021}. Consequently, we have chosen an epidemic model based on ordinary differential equations (\textit{ODEs}) that has already been validated in~\cite{kuhn_assessment_2021,koslow_appropriate_2022} and which categorizes subpopulations based on the infection states \textit{susceptible}, \textit{exposed (not infectious)}, \textit{a- or presymptomatic and infectious}, \textit{mildly symptomatic and infectious}, \textit{severely infected (hospitalized)}, \textit{critically infected (ICU)}, \textit{dead}, and \textit{recovered}; see~\cref{fig:InfectionStates}. Aside the mentioned indicators, the model particularly considers pre- and asymptomatic transmission as studies on SARS-CoV\nobreakdash-2 observed that viral load is peaking close to symptom onset as observed by \MK{Walsh et al. and Puhach et al.}~\cite{walsh_duration_2020,puhach_sars-cov-2_2023}. The model is efficiently implemented in C++ in the MEmilio~\cite{Bicker_MEmilio_2026} framework, which means that it yields very fast model evaluations on a single core. 
We furthermore divided the population into $G=6$ age groups, that is, from 0 to 4, 5 to 14, 15 to 34, 35 to 59, 60 to 79, and 80+ years. \MK{While also simpler models of \textit{SIR}-type structure and without age-structure could be learnt by a surrogate model, the resulting surrogate model could then not be evaluated for critical numbers on hospitalizations, intensive care, or deaths and without learning about most contributing age groups. In addition, the \textit{SIR}-type structure would miss the well observed disease evolution of COVID-19 with exposed, non-infectious and nonsymptomatic, infectious states.}

Transmission is modeled via the susceptible population $S_i(t)$ becoming exposed ($E_i(t)$), through infection of either infectious nonsymptomatic ($I_{NS,i}(t)$) or symptomatic individuals ($I_{Sy,i}(t))$, $i \in \{1,\dots,6\}$. This can be described by the ODE
\begin{align}\label{eq:susceptible}
    \frac{d S_{i}(t)}{d t} &= - S_{i}(t)\rho_{i}\,\sum_{j=1}^6 \phi_{i,j}(t) \frac{\xi_{I_{NS},j} I_{NS,j}(t) + \xi_{I_{Sy},j} I_{Sy,j}(t)}{N_j-D_j(t)},
\end{align}
where $\xi_{I_{NS}}$ and $\xi_{I_{Sy}}$ denote the proportion of nonisolated a- or presymptomatic and symptomatic infectious individuals, respectively. Furthermore, $\rho_i$ denotes the age-resolved probability of becoming infected from a contact with an infectious person, and $\phi_{i,j}(t)$ represents the mean daily number of contacts between individuals of age group $i$ and age group $j$. While, theoretically, $\phi_{i,j}(t)$ could vary arbitrarily, in our realization, it only varies in the context
of a contact change point and is constant otherwise; cf.~\cref{eq:phi}. Additionally, we denote the initial population in age group $j$ by $N_j$ and the number of disease-related deaths in age group $j$ up to time $t$ by $D_j(t)$, $j=1,\ldots,6$.

Disease progression follows a chain of compartments, where an individual either recovers or moves to an aggravated state. We define $z_{j,i}$ as the $j$-th disease state for age group $i$ and denote the probability of transition from state $z_{j,i}$ to aggravated state $z_{j+1,i}$ by $\mu_{z_{j,i}}^{z_{j+1,i}}$, while consequently $1 - \mu_{z_{j,i}}^{z_{j+1,i}}$ is the probability of recovery. The average time an individual spends in a given state $z_{j,i}$ is represented by $T_{z_{j,i}}$. For disease states where progression or recovery is possible, i.e., excluding \textit{susceptible}, \textit{exposed (not infectious)}, \textit{recovered}, and \textit{dead}, transition follows the form
\begin{align}\label{eq:exceptsusceptible}
    \frac{d z_{j,i}(t)}{d t} &= \mu_{z_{j-1,i}}^{z_{j,i}}\frac{ z_{j-1,i}(t)}{T_{z_{j-1,i}}}-\frac{z_{j,i}(t)}{T_{z_{j,i}}},
\end{align}
where $z_{j-1,i}$ is the former, less severe state.
For the full set of equations, we refer to~\cite{kuhn_assessment_2021}. 

In this study, we use parameters that were obtained for the first waves of wild-type SARS-CoV\nobreakdash-2 in 2020. Identical to \MK{Plötzke et al.}~\cite{PLOTZKE2026823}, the parameters are either directly obtained or calculated based on~\cite[Table~2]{kuhn_assessment_2021}. 
For the average values over all age groups, the age-specific values are weighted with respect to share of the corresponding age group in the total population provided for Germany by~\cite{regionaldatenbank_deutschland_fortschreibung}. The overview on the particular values can be found in~\cref{tab:model_parameters,tab:COVID-19parameters}.

%  0–4 5–14 15–34 35–59 60–79 80+ 

\begin{figure} 
\begin{center}
\begin{minipage}{0.65\textwidth}
 \begin{tikzpicture}[node distance=4cm, every node/.style={scale=0.75}]
\tikzstyle{block} = [rectangle, minimum width=3cm, minimum height=1cm, text centered, draw=black, line width=0.5mm, fill=gray!30]
\tikzstyle{arrow} = [thick,->,>=stealth, line width=0.5mm]
% Erste Reihe
\node (susceptible) [block, fill={rgb,255:red,176; green,176; blue,176}] {\color{white}\textbf{Susceptible}}; % Grau
\node (exposed) [block, fill={rgb,155:red,217; green,232; blue,252}, right of=susceptible] {\color{white}\textbf{Exposed}}; % Hellblau
\node (no_symptoms) [block, fill={rgb,255:red,245; green,134; blue,52}, right of=exposed] {\color{white}\textbf{A-/Presympt.}}; % Orange
\node (mild) [block, fill={rgb,255:red,192; green,0; blue,0}, below of=no_symptoms, yshift=1cm] {\color{white}\textbf{Mild}}; % Rot

% Spalten
\node (severe) [block, fill={rgb,255:red,204; green,51; blue,153}, below of=mild, yshift=1cm] {\color{white}\textbf{Severe}}; % Magenta
\node (critical) [block, fill={rgb,255:red,112; green,68; blue,160}, left of=severe] {\color{white}\textbf{Critical}}; % Lila

\node (recovered) [block, fill={rgb,255:red,40; green,167; blue,69}, below of=exposed, yshift=1cm] {\color{white}\textbf{Recovered}}; % Grün
\node (dead) [block, fill={rgb,255:red,77; green,77; blue,77}, left of=critical] {\color{white}\textbf{Dead}}; % Dunkelgrau

% Pfeile
\draw [arrow] (susceptible) -- (exposed);
\draw [arrow] (exposed) -- (no_symptoms);
\draw [arrow] (no_symptoms) -- (mild);
\draw [arrow] (mild) -- (severe);
\draw [arrow] (severe) -- (critical);

% Verbindungen zu den Endzuständen
\draw [arrow] (no_symptoms) -- (recovered);
\draw [arrow] (mild) -- (recovered);
\draw [arrow] (severe) -- (recovered);
\draw [arrow] (critical) -- (recovered);
\draw [arrow] (critical) -- (dead);
\end{tikzpicture}
\end{minipage}
	\caption{\textbf{Disease state transition model.} The disease state transition model comprises eight infection states with possibility for recovery or aggravation after each disease state following exposure state (except for death and recovery itself).}
	\label{fig:InfectionStates}
\end{center}
\end{figure}

\begin{table}[!h]
\centering
\begin{tabular}{ll}
\textbf{Parameter}  & \textbf{Description} \\ \hline \hline
\texttt{$T_E$}  & Duration (in days) in \textit{exposed (not infectious)} state \\ \hline 
\texttt{$T_{INS}$}  & Duration (in days) in \textit{a- or presymptomatic and infectious} state \\ \hline
\texttt{$T_{Sy}$}  & Duration (in days) in \textit{mildly symptomatic and infectious} state \\ \hline
\texttt{$T_{Sev}$}   & Duration (in days) in \textit{severely infected (hospitalized)} state \\ \hline
\texttt{$T_{Cr}$}  & Duration (in days) in \textit{critically infected (ICU)} state \\ \hline
\texttt{$\rho$}  & Transmission risk on contact\\ \hline
\texttt{$\xi_{I_{NS}}$}  & Proportion of not isolated a- or presymptomatic infectious cases \\ \hline
\texttt{$\xi_{I_{Sy}}$}  & Proportion of not isolated symptomatic cases \\ \hline
\texttt{$\mu_{I_{NS}}^{I_{Sy}}$}  & Proportion of symptomatic cases per all a- and presymptomatic 
 infectious cases \\ \hline
\texttt{$\mu_{I_{Sy}}^{I_{Sev}}$}  & Proportion of severe cases per symptomatic case\\ \hline
\texttt{$\mu_{I_{Sev}}^{I_{Cr}}$}  & Proportion of critical cases per severe case\\ \hline
\texttt{$\mu_{I_{Cr}}^{D}$}  & Proportion of deaths per critical case \\
\end{tabular}
\caption{\textbf{Explanation of the model parameters.} }
\label{tab:model_parameters}
\end{table}

\begin{table}
    \centering
\def\arraystretch{1.25}
\begin{tabular}{cccccccc}
      &  \textbf{0--4~y} & \textbf{5--14~y} & \textbf{15--34~y} & \textbf{35--59~y} & \textbf{60--79~y} & \textbf{80+~y} & \textbf{Weighted Average} \\
\hline \hline
    $T_{E,i}$ & $3.335$ & $3.335$ & $3.335$ & $3.335$ & $3.335$ & $3.335$ & $3.335$\\
     $T_{I_{INS},i}$ & $2.74$ & $2.74$ & $2.565$ & $2.565$ & $2.565$ & $2.565$ & $2.58916$\\
      $T_{I_{Sy},i}$ & $7.02625$ & $7.02625$ & $7.0665$ & $6.9385$ & $6.835$ & $6.775$ & $6.94547$\\
    $T_{Sev}$ & $5$ & $5$ & $5.925$ & $7.55$ & $8.5$ & $11$ & $7.28196$\\
     $T_{Cr,i}$ & $6.95$ & $6.95$ & $6.86$ & $17.36$ & $17.1$ & $11.6$ & $13.066$ \\
     \midrule
 $\rho_i(t)$ & $0.03$ & $0.06$ & $0.06$ & $0.06$ & $0.09$ & $0.175$ & $0.07333$\\
\midrule
   $\mu_{I_{NS}}^{I_{Sy}}$ & $0.75$ & $0.75$ & $0.8$ & $0.8$ & $0.8$ & $0.8$ & $ 0.79310$\\
     $\mu_{I_{Sy}}^{I_{Sev}}$ & $0.0075$ & $0.0075$ & $0.019$ & $0.0615$ & $0.165$ & $0.225$ & $0.07864$\\
     $\mu_{I_{Sev}}^{I_{Cr}}$ & $0.075$ & $0.075$ & $0.075$  & $0.15$ & $0.3$ & $0.4$ & $0.17318$\\
     $\mu_{I_{Cr}}^{D}$ & $0.05$ & $0.05$ & $0.14$ & $0.14$ & $0.4$ & $0.6$ & $0.21718$\\     
\bottomrule
\end{tabular}
\caption{\textbf{Age-resolved parameters for wild-type SARS-CoV\protect\nobreakdash-2 based on~\cite{kuhn_assessment_2021}.}}
\label{tab:COVID-19parameters}
\end{table}

Contact pattern changes are modeled within the contact rate matrix $\phi (t) =(\phi_{i,j}(t))_{i,j}$. Without loss of generality, let us assume that the first contact change takes place at $c_1\in\mathbb{R}$ with $c_1 > c_0$, where $c_0$ is the initial time. 
For a homogeneous reduction factor $0\leq r\leq1$, we define
\begin{align}\label{eq:phi}
{\phi}_{i,j}(t) := 
    \begin{cases}
        \phi_{i,j,0},              & t \leq c_1 \\
        \widehat{\phi}_{i,j}(t), & t \in (c_1, c_1 + \delta) \\
        (1-r)\phi_{i,j,0} , & t \geq c_1 + \delta
    \end{cases}, \quad 0<\delta< 1,
\end{align}
where $\phi_{i,j,0}$ is the initial contact rate, $\delta$ defines a transition interval, and $\widehat{\phi}_{i,j}$ represents a transition contact rate to ensure smooth transition, i.e., ${\phi}_{i,j}(t)\in\mathcal{C}^1((c_1-\varepsilon,c_1+ \delta + \varepsilon))$ for ${i,j}\in\{1,\ldots,6\}$ with $0<\varepsilon\ll1$. As suggested in~\cite{kuhn_assessment_2021}, $\widehat{\phi}$ is realized through a cosine function.

In order to realize the spatial resolution, we employed a generalized graph as described in~\cite{kuhn_assessment_2021}. Proceeding with this approach, we obtained a graph with one node for each spatial region, which is then equipped with a a local disease dynamics model based on ODEs. Furthermore, we obtained one edge for each pair of age group and disease state, which then implements the corresponding mobility exchange. The $400\times 400$ \textit{mobility matrix} for all of Germany's counties was extrapolated from the German Federal Employment Agency~\cite{bmas_pendlerverflechtungen_2020}. The adjacency matrix $A$ needed for the GNN was obtained from this mobility data matrix by setting all nonzero connections to one.

\MK{The resulting graph, representing 400 German counties, is directed and has a density of $30.4\%$ with $48,467$ edges. Its topology is characterized by an uneven degree distribution with highly connected hubs (e.g., Frankfurt) connecting to smaller surrounding areas, indicated by a slightly negative assortativity coefficient ($-0.096$). A high clustering coefficient ($0.56$) further reflects strong local community structures. From a physical perspective, the GNN learns to approximate the epidemic diffusion process on this specific topology. Since the local rules of this diffusion are consistent, the model is capable of generalizing to unseen temporal scenarios, given the local connectivity patterns remain similar.}

While also this metapopulation model was implemented efficiently and did undergo several optimization steps, for large experiments we still require relevant computing resources; cf.~\Cref{fig:GNN_results}, Panel C.

% \begin{figure} 
% \begin{center}
% \begin{minipage}{0.8\textwidth}
% \includegraphics[width=\textwidth]{boxplot_input_compartments_noagegroups.png}
% \end{minipage}
% 	\caption{\textbf{Sampled variation of initial conditions for the different compartments in the ODE model.}
%     %\tdAH{Make the $y$ axes labels consistent.}
%     }
% 	\label{fig:initial_ode}
% \end{center}
% \end{figure}

\subsection{Performance evaluation and objective function}

\MK{Simulating the previously described mechanistic metapopulation model}, we obtain time-dependent (and spatially resolved) outputs. The output vectors from all \MK{local} models \MK{of the form~\cref{eq:susceptible,eq:exceptsusceptible}} are interpolated to full simulation days so that corresponding \MK{age-structured} and spatially resolved data sets have the same dimension. To evaluate the prediction performance of the neural network-based surrogate models as well as the dummy estimator (see next subsection), we use the mean absolute percentage error (MAPE) as evaluation metric, which is defined as
\begin{align}\label{eq:mape_def}
\textnormal{MAPE}:=\frac{100}{n} \sum_{i=1}^n \frac{|y_{i}-\widehat{y}_{i}|}{|y_{i}|},
\end{align}
as objective function. Here, $n$ is the \MK{total} number of \MK{data points} \MK{which is the product of the number of runs, the number of days, and the number of compartments}, $y_{i}$ \MK{and $\widehat{y}_{i}$ denote individual scalar values representing} the interpolated "ground truth" \MK{and predicted daily values, respectively}

\subsection{Generation of input data} 

To develop a generalizable surrogate model for a large variety of observed disease dynamics, we require heterogeneous data sets with a large variance in the outcomes, which transfer to the input features of our surrogate models and which eventually cover the full range of all relevant scenarios. The heterogeneity of the generated data sets were assessed qualitatively by visualization and quantitatively by investigating the predictive power of a simple \textit{dummy estimator}. We compute this dummy estimator by the averaging over 80\,\% of the data and evaluating its predictive performance on the trajectories of the remaining 20\,\% of the data using the MAPE given in Eq.~\eqref{eq:mape_def} as the performance metric. We conclude that the higher the MAPE of the dummy estimator, the higher the variance of the input data. 

\begin{figure}[!h]
  \centering
  \includegraphics[width=\linewidth]{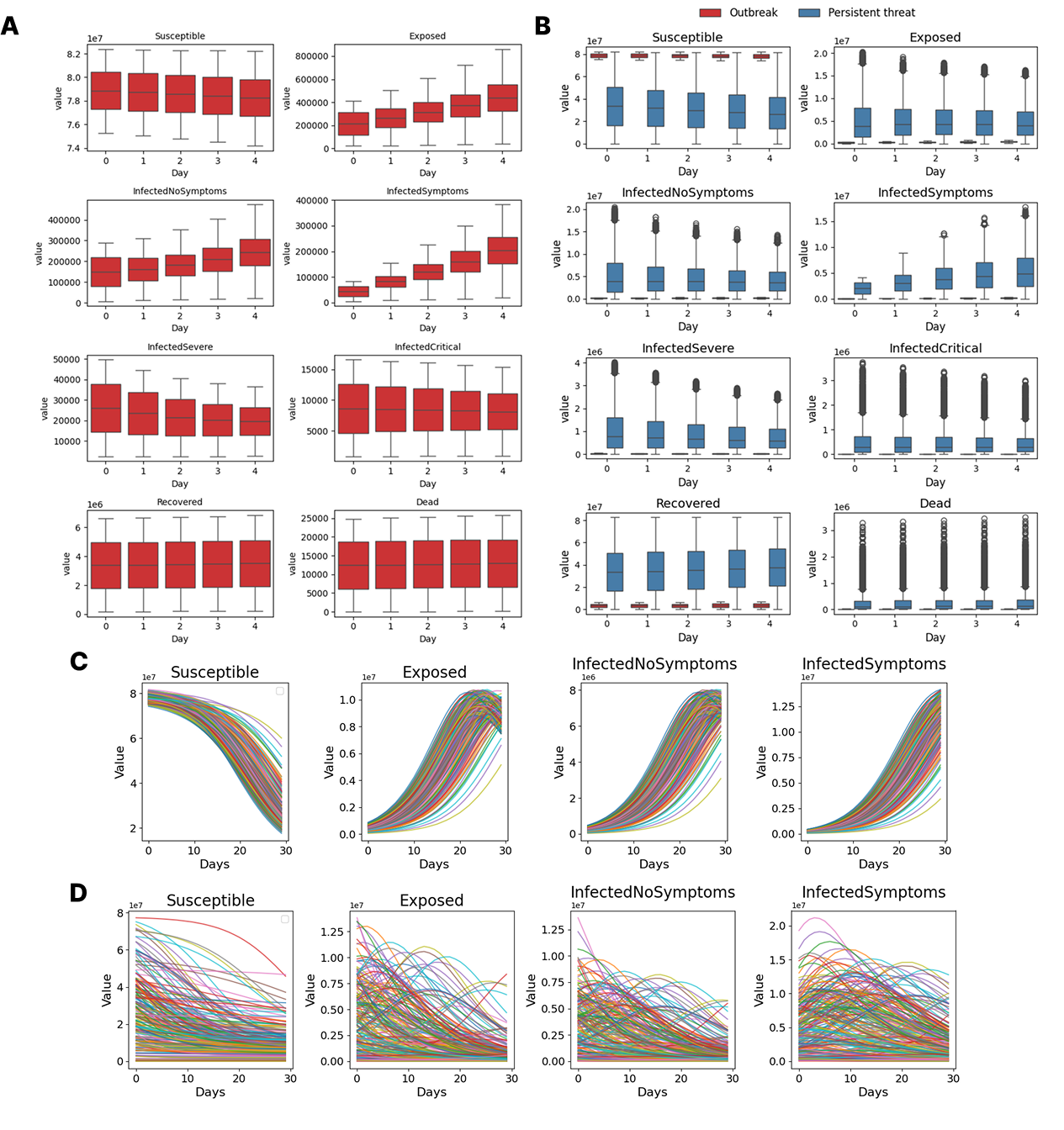}
  \caption{\textbf{Initial conditions over five input days and trajectories over the prediction horizon.} A) Initial conditions for outbreak range scenarios. B) Comparison between initial conditions for outbreak range and persistent-threat range scenarios. C) The two hundred first trajectories for the outbreak scenarios. D) The two hundred first trajectories for the persistent-threat scenarios.  }
  \label{fig:input_trajectories_boxplots}
\end{figure}

We consider two epidemic regimes that define the initial five input days per sample (values per 100\,000 population and stratified by six age groups unless stated otherwise). In the \emph{outbreak} regime, currently symptomatic individuals are sampled uniformly from $[7,100]$ and recently exposed from $[25,500]$; the remaining compartments are filled consistently to satisfy population balance. In the \emph{persistent-threat} regime, the share of currently symptomatic individuals is drawn uniformly from $[0.01\%,5\%]$; in the worst case, up to $50\%$ of the population may be recently exposed, a-/presymptomatic, or symptomatic. To reflect heterogeneous historical dynamics and immunity, the initial fraction recovered per region is drawn uniformly within the feasible interval implied by the ongoing dynamics so that all compartment shares remain nonnegative and sum to one. The variability of the outbreak and persistent-threat ranges are provided in Panels A and B of~\Cref{fig:input_trajectories_boxplots}. In Panels C) and D) of~\Cref{fig:input_trajectories_boxplots}, we plot the 200 first trajectories for the outbreak scenario (Panel C) and the persistent-threat (Panel D) settings.
Here, we see a much more heterogeneous development in Panel D).

Interventions are represented by contact change points. Each sample includes $M\in\{0,1,2,3\}$ changes within the first 30 days; $M$ is drawn uniformly from this set. Change days $(d_1,\dots,d_M)$ are drawn without replacement from $\{1,\dots,30\}$ and sorted increasingly, and the homogeneous reduction factors $r_m$ are sampled from $\mathcal{U}[0,1)$. Smooth transitions over a short interval $\delta\in(0,1)$ are applied in the mechanistic simulator using a cosine ramp (cf.~\cref{eq:phi}). Age-structured contact matrices are $6\times6$ per change point; if $M<3$, the unused slots are masked in the learning input.

We model Germany as a graph with 400 nodes (counties). Each node carries the age-stratified compartment trajectories. The adjacency matrix $A\in\{0,1\}^{400\times400}$ is derived from the afore-mentioned mobility matrix by mapping nonzero flows to one; this yields a sparse graph with roughly $25\%$ nonzero entries. The same graph is used for training and testing.

\begin{figure}[!t]
  \centering
  \includegraphics[width=\linewidth]{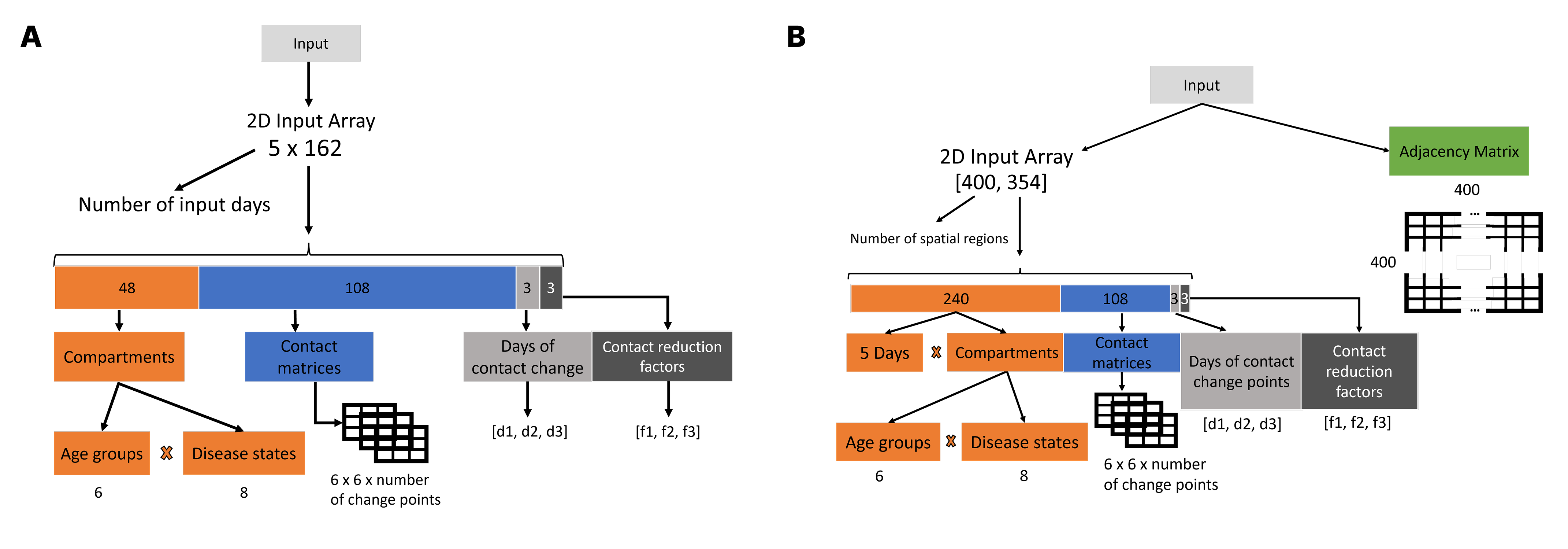}
  \caption{\textbf{Non-spatial vs.\ spatial input encoding.}
  A) Non-spatial 2D input (5\,$\times$\,162): five input days with 48 (age groups $\times$ compartments) values
  plus up to three $6\times6$ contact matrices, their change days, and reduction factors.
  B) Spatial GNN input: for each node (400 counties) the same features as in Panel A are broadcast and paired with the adjacency matrix.}
  \label{fig:input_structure}
\end{figure}

Per sample we provide the first five days as input and predict subsequent horizons of 30, 60, or 90 days. For non-spatial baselines, we use a 2D array of dimension $5\times162$ array with the five input days defining the rows each with $48$ age-resolved compartment values (six age groups times eight compartments) plus up to three $6\times6$ contact matrices with their three contact change days and reduction factors. For the GNNs, each sample consists of node features $X\in\mathbb{R}^{400\times354}$ containing the five input days of compartment values ($48$ per node) and the intervention descriptors (contact matrices, change days, reduction factors) broadcast per node, paired with the adjacency matrix $A$. This schema mirrors the input layout illustrated in~\cref{fig:input_structure}. To balance scale differences across compartments (e.g., \emph{Susceptible} vs.\ ICU), we apply a $\log(1+x)$ transform to all count-valued series before training; evaluation is performed on the original scale after inverse transforming.

We generate $10{,}000$ samples for the non-spatial baselines and $1{,}000$ samples for the spatial GNNs. Unless stated otherwise, we use an 80–10–10 split into train/validation/test on the sample level and perform five-fold cross-validation on the training split during architecture search to avoid split artifacts; the reported test MAPE refers to the held-out test set. The dummy estimator is computed on the training split and evaluated on the held-out portion within each regime as a single-number proxy for input heterogeneity; detailed values are reported in the Results.

\subsection{Neural network surrogate model} 

We conduct a multi-stage model building and optimization process to build up a neural network with spatially resolved infection predictions. Before deriving suitable input data structures for learning a spatially resolved graph-based model with a GNN, we consider the temporal evolution of disease dynamics. For this, we train an MLP, an LSTM and a CNN model, each implemented using the Keras~\cite{chollet_keras_2015} library with TensorFlow~\cite{tensorflow2015} as backend.

With respect to the temporal evaluation, we take several substeps in our tests. In particular, we first consider a population without stratification into age groups. Then, we stratify the population into age groups, for which we consider age-dependent contact patterns and disease parameters. For each model architecture, we perform a grid search on zero to four hidden layers 
with 32 to 1024 neurons per layer, ranging up to seven layers and 2048 neurons for the final GNN layer type. After selecting a suitable neural network architecture, we fine-tune additional architecture-specific hyperparameters. 

After our preliminary tests with simple neural network models which do not consider spatial effects, we turn to GNNs. We use the Spektral library \MK{from Grattarola and Alippi}~\cite{grattarola_graph_2021} for defining the graph data and building the GNNs and Keras to train and evaluate the model. 
Spektral is an open-source Python library based on the Keras API and TensorFlow 2~\cite{tensorflow2015} implementing GNN models. Other existing libraries, which are not considered here, are PyTorch Geometric~\cite{fey_fast_2019} and DeepGraphLibrary~\cite{wang_deep_2020}.

GNNs can be categorized into spectral and spatial approaches, where spectral approaches have a mathematical foundation in graph signal processing while spatial approaches directly aggregate information from the nodes' spatial neighborhoods (\MK{see Wu et al.}~\cite{wu_comprehensive_2021}), which is given by the graph. We compare the performance of four different types of GNN layers available in Spektral to build our model. As a first approach, we consider GCNConv layers, which are based on the graph convolutional network (\textit{GCN}) as proposed by \MK{Kipf et al.}~\cite{kipf_semi-supervised_2017}. In this spectral approach, a modified adjacency matrix with self-loops is used. The second type are ARMAConv layers, which are based on \MK{Bianchi et al.}~\cite{bianchi_graph_2021}. Like the GCNConv layer, this is a spectral approach, and it applies convolutional operations in combination with an autoregressive moving average (\textit{ARMA}) approach for improved training performance; cf.~the discussion below. The third approach uses approximative personalized propagation of neural predictions convolutional (\textit{APPNPConv}) layers based on \MK{Gasteiger et al.}~\cite{gasteiger_predict_2022}, which can also be categorized as a spectral approach with convolutional operations. Lastly, we consider the graph attention convolutional (\textit{GATConv}) layer from \MK{Velickovic et al.}~\cite{velickovic_graph_2018}, which is different from the other considered approaches, in the sense that it uses a spatial approach that implements an attention mechanism.

Since the ARMAConv layer performed best (see the \textit{Results} section), we briefly provide some details on this specific type of GNN layer as well as our final architecture. In~\cite{bianchi_graph_2021}, \MK{Bianchi et al.} substituted polynomial filters in GNNs by filters which are based on the ARMA approach.
In particular, in our optimized architecture, the final model consists of seven ARMAConv layers with 512 channels per layer and consisting of one stack of GCS layers, which serves as a filter on the node feature input data. Generally, we denote a ARMAConv layer with $k$ GCS stacks as $\textit{ARMA}_k$.
The $i$th GCS layer in the stack is parametrized by the trainable weight matrix $\textbf{W}_i$ as well as the trainable skip connection $\textbf{V}_i$, which arises from the ARMA scheme applied to the GCS stack. In the case of the $\textit{ARMA}_1$ layer with one GCS stack, the graph convolution layer corresponds to
\begin{align*}    
\overline{\textbf{X}}_1^{(t+1)} = \sigma(\tilde{\textbf{A}}\overline{\textbf{X}}_1^{(t)}\textbf{W}_1^{(t)}+ \overline{\textbf{X}}^{(0)} \textbf{V}_1^{(t)}),
\end{align*}
where $\sigma$ is an activation function, $\overline{\textbf{X}}^{(0)}$ are the initial node features, $\overline{\textbf{X}}^{(t)}$ are filtered node features of the previous iteration, and $\tilde{\textbf{A}}$ is the normalized adjacency matrix
$\tilde{\textbf{A}} = \textbf{D}^{-1/2}\textbf{A}\textbf{D}^{-1/2}$; it is normalized using the degree matrix $\textbf{D}$, which counts for each node the number of adjacent edges in the graph. 
After applying all GCS layers, a pooling operations is employed on the GCS stack, for instance, average pooling.
With only one GCS layer, the averaging pool does not have any effect and simply applies the identity. In all configurations, the activation function was set to ReLU and no dropout was selected. 

\section{Results}
\label{sec:results}

In this section, we present the results of our computational experiments. First, we compute the baseline performance to be beaten by any meaningful neural network. Then, we discuss how we optimize the architecture and hyperparameters of our models, starting with simple models accounting only for the temporal evolution and moving on to models also taking into account the spatial distribution. Finally, we show results on the computational efficiency of the models.

\subsection{Baseline performance}

The design of experiments in~\Cref{sec:methods} produces substantially different trajectories across, both, the \emph{outbreak} and \emph{persistent‑threat} scenarios, with differently large variances in the data set. The representative ranges and trajectories are shown in \Cref{fig:input_trajectories_boxplots}. As mentioned in the previous section, we use the dummy estimator to compute the baseline performance to be beaten by any meaningful surrogate model. As a compact quantitative proxy for dataset variability, the mean dummy estimator (trained on 80\,\% of the data and evaluated on the remaining 20\,\%) attains MAPE values of 34.16\,\% for the outbreak regime and 1197.76\,\% for the persistent‑threat regime (original scale). The latter highlights a markedly harder prediction setting and thus a useful stress test for model selection.

\subsection{Neural network performance without spatial resolution}
In this section, we consider only temporal evolution and neglect spatial effects to get a better idea of the potential performances on the complex datasets generated and compare the model failures against the dummy estimator outcomes of the previous section. 

\begin{figure}[!h]
\begin{adjustwidth}{-.35in}{-.35in}
    \includegraphics[width=1.1\textwidth]{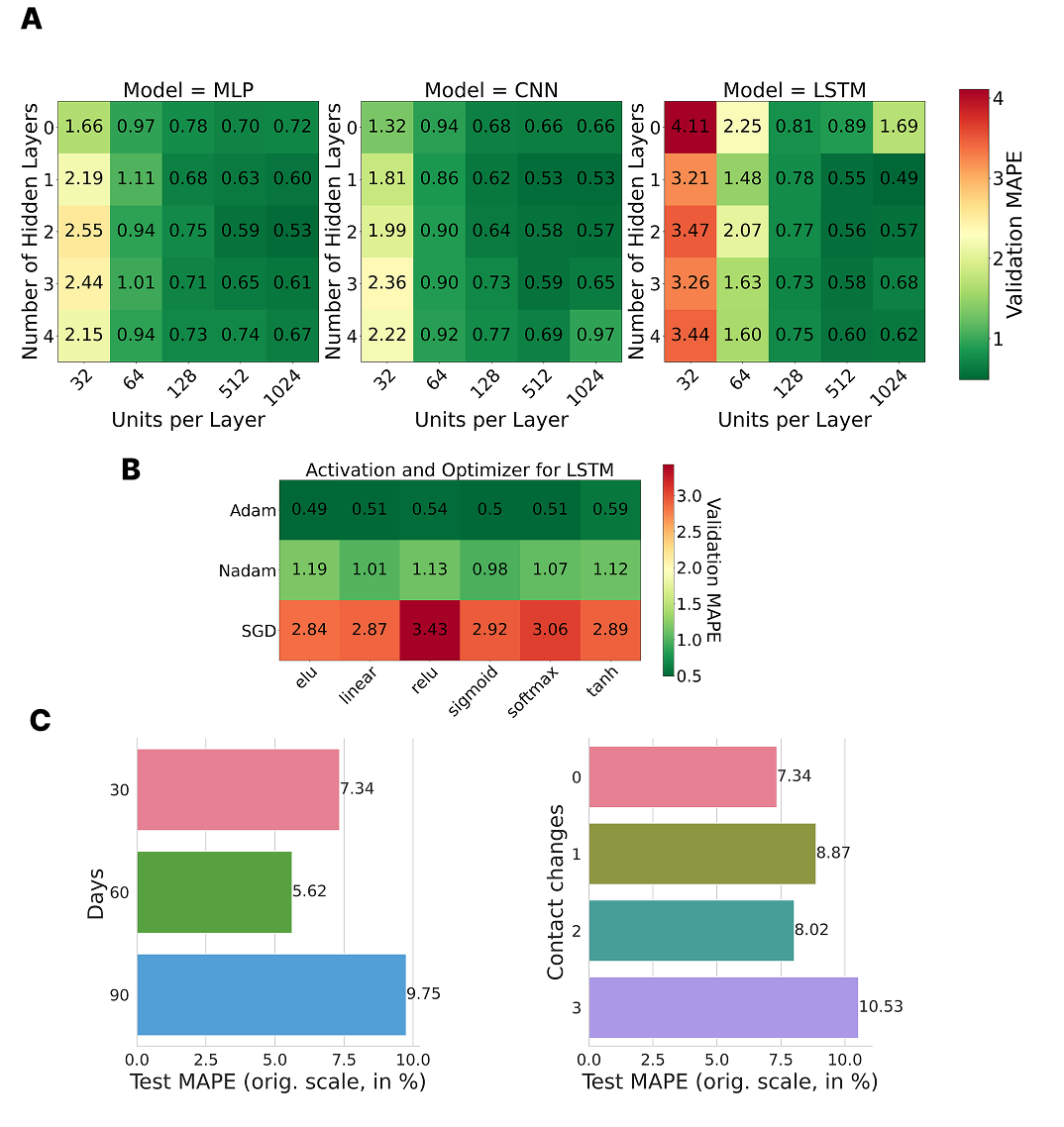}
	\caption{\textbf{MAPE results of the grid search for the simple neural networks and simple input data structure.} A) Validation MAPE results (in log-scale) for a grid search for simple models with six age groups and no contact change on a horizon over 30 days. B) Validation MAPE results (in log-scale) for the grid search for different optimizers and activation functions. C) Test MAPE results (in original scale) for the final LSTM model on different prediction horizons (30, 60, and 90 days) and for different numbers of contact changes (0, 1, 2, 3 changes in 30 days).
    }
	\label{fig:nonspatialresults}
\end{adjustwidth}
\end{figure}

We consider simple neural networks, such as multilayer perceptrons (\textit{MLPs}), convolutional neural networks (\textit{CNNs}) with one-dimensional convolutions in the time dimension, and long short-term memory (\textit{LSTM}) networks. 
We conduct grid searches on the number of hidden layers (0 to 4) and neurons (neurons, channels, or units) per layer (32 to 1024) for a setting with six age groups and simulated over 30 days without contact change points. We used an 80-10-10 split of the dataset into training, validation, and test sets.
All grid searches are conducted by using a five-fold cross validation (CV) to avoid artifacts from the data split. The logarithmically scaled MAPE values of the cross-validation are shown in~\Cref{fig:nonspatialresults} (Panel A). The MAPE on the validation sets will be denoted as \textit{validation MAPE}, the MAPE on a corresponding test set will be denoted as \textit{test MAPE}. A further analysis revealed Adam as the best optimizer and \MK{ReLU} as the best activation function; see~\Cref{fig:nonspatialresults} (Panel B). While we evaluated the models on both datasets, outbreak and persistent-threat, we only provide the latter case, which is the substantially more complex one. The best architecture in our comparison was a simple LSTM with one hidden layer and 1024 units which achieved a MAPE of 0.49~\% across the grid search and, thus, 3-4 magnitudes better than our dummy baseline estimator. We further test it on longer time horizons with up to three contact change points; see~\Cref{fig:nonspatialresults} (Panel C). While the architecture is optimized on a much simpler model, we obtain test MAPEs between only 5-10\,\% for the input features depicted in~\Cref{fig:input_structure} (Panel A). In the supplementary material, \cref{fig:nns_trajectories_1,fig:nns_trajectories_2}, we provide predictions for the first two trajectories with and without stratification into age groups to visually assess the meaning of the corresponding MAPE values.

\begin{figure}[!h]
\begin{adjustwidth}{-.35in}{-.35in}
\includegraphics[width=1.1\textwidth]{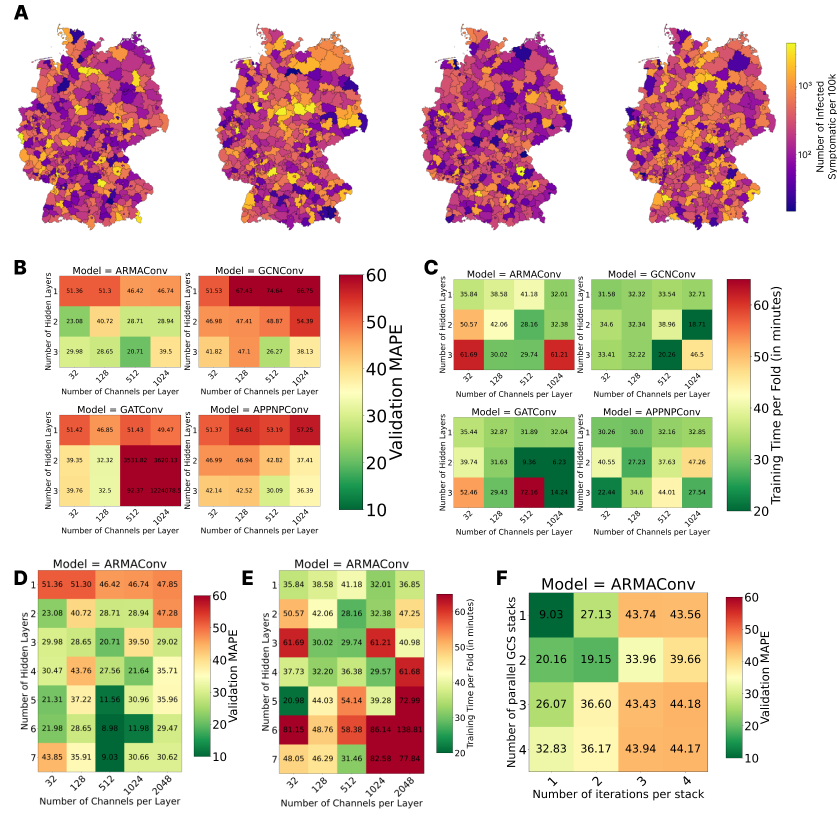}
\caption{\textbf{Spatially distributed disease dynamics and grid search of different GNN infrastructures.} A) Four random initializations of heterogeneous disease dynamics over the 400 counties of Germany. B) Validation MAPEs \MK{(in log-scale)} for the four different GNN layers. C) Training times corresponding to Panel B). D) Extended grid search for ARMAConv model \MK{(in log-scale)}. E) Training times corresponding to Panel D). F) Further hyperparameter tuning for ARMAConv model with seven layers and 512 channels per layer \MK{(in log-scale)}.
}
\label{fig:spatial_dynamics_GNN_gridheat}
\end{adjustwidth}
\end{figure}

\subsection{Optimized GNN architecture}

Combining the results from our explorational experiments of the simple MLP and LSTM neural networks with some minor adaptations\MK{, i.e., eliminating the redundant storage of the contact matrices of the 2D dimensional $5\times 162$ and transforming it to a one dimensional vector of length 354}, we use the input data set structure for the GNNs as presented in the previous section and shown in~\cref{fig:input_structure} (Panel B). For a GNN to be trained or executed, we provide an (additional) adjacency matrix $A$ for the considered regions.
The adjacency matrix is naturally obtained from the graph-based expert modeling approach. The dimension of the adjacency matrix is $400\times 400$, which comes from the particular application on the 400 German counties. Similarly, the first array dimension of the remaining input is 400. Eventually, to adequately integrate heterogeneous disease dynamics over an entire country, we randomize historic and ongoing disease dynamics over all 400 counties. 
The ongoing disease dynamics are initialized per county with the persistent-threat scenario, thus allowing between very small and very high infectious disease dynamics. For the first four initializations, see~\Cref{fig:spatial_dynamics_GNN_gridheat} (Panel A). To additionally randomize for the historic dynamics and immunities, we draw the number of recoveries uniformly from the interval between zero and one minus the maximum ongoing dynamics (infected and dead) that are possible in the persistent-threat range.

Having derived a suitable set of input features for a GNN surrogate model, we selected an appropriate GNN model and optimized its architecture and hyperparameters. We conducted a grid search to identify the model architecture with the best performance, also taking into account the respective training time, which can be considerable. To avoid unnecessarily long training times, we used an early stopping criterion after 200 iterations without improvement (i.e., patience of 200 iterations). \MK{Again, averaged validation MAPEs are presented in log-scale.}

We start by considering different graph layers as described in the \textit{Materials and Methods} section and vary the number of layers between 1 and 3 and the number of channels per layer from 32 to 1024. We observe that the ARMAConv model with three layers and 512 channels per layer performs best with respect to the validation MAPE (\Cref{fig:spatial_dynamics_GNN_gridheat}, Panel B), while only the GATConv models with the highest MAPEs need substantially less training time (\Cref{fig:spatial_dynamics_GNN_gridheat}, Panel C). We thus refine the grid search for the ARMAConv model to seven layers and observe that the models with six or seven layers and 512 channels per layer perform best. While the validation MAPE for seven layers, 9.03~\%, is slightly larger than that for six layers, 8.98~\% (\Cref{fig:spatial_dynamics_GNN_gridheat}, Panel D), the training time for seven layers, 31.46 minutes, is reduced by almost a factor of two compared with the model with six layers, which requires 58.38 minutes of training (\Cref{fig:spatial_dynamics_GNN_gridheat}, Panel E). \MK{This counterintuitive reduction in total training time results from the early stopping mechanism: although the computational cost per epoch increases with model depth, the 7-layer model converged in substantially fewer epochs across the cross-validation folds compared to the 6-layer model.}
\MK{To ensure that this large architecture remains stable, we analyze the learning dynamics. As shown in~\cref{fig:learning_curve}, training and validation errors decrease without divergence. At the best epoch (1277), the validation MAPE is comparable to the training MAPE, confirming robust generalization without overfitting.
\begin{figure}[h]
    \centering
    \includegraphics[width=0.6\linewidth]{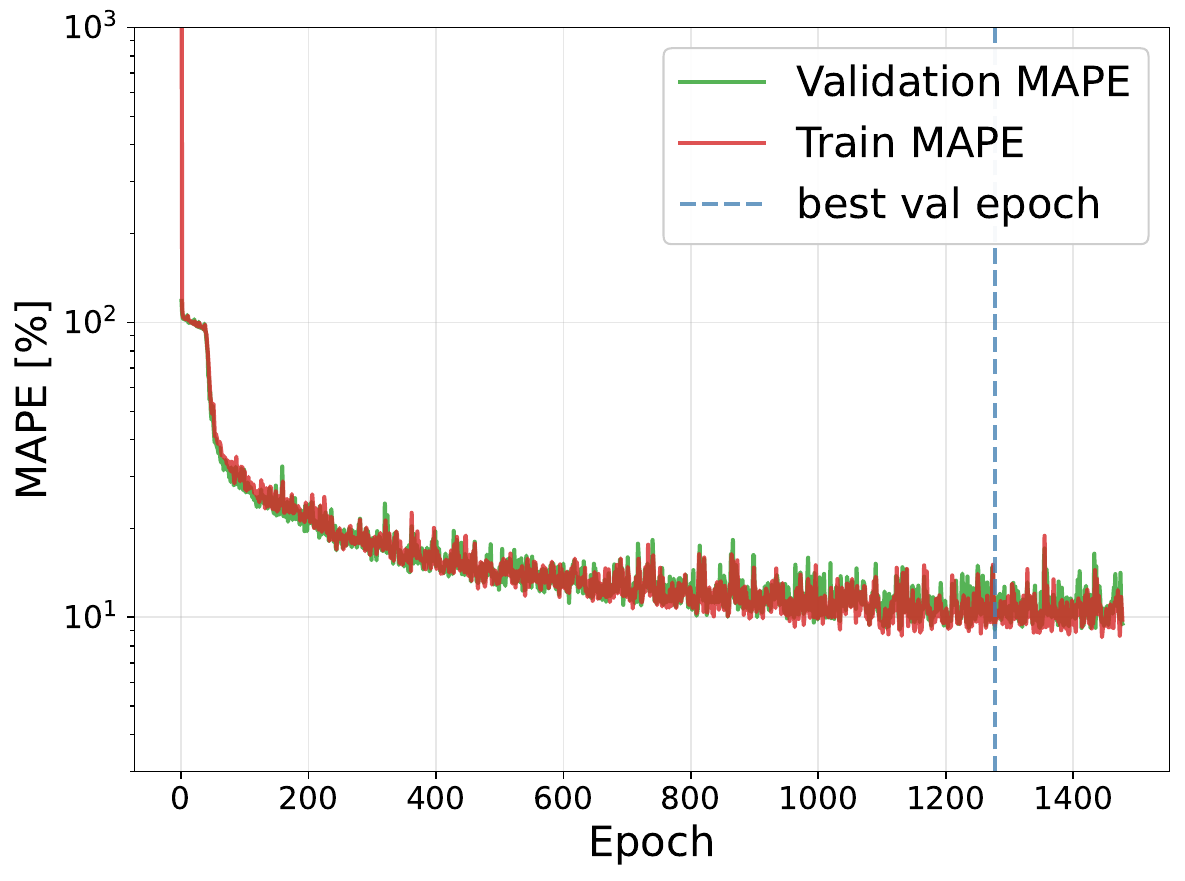}
    \caption{\MK{\textbf{Learning dynamics of the optimized ARMAConv model (7 layers, 512 channels).} The training (red) and validation (green) MAPEs show consistent convergence.}}
    \label{fig:learning_curve}
\end{figure}
}

In~\Cref{fig:spatial_dynamics_GNN_gridheat} (Panel F), we report an additional assessment on hyperparameter tuning, testing the parameters that are specific to the ARMAConv model: the numbers of parallel GCS stacks and iterations per stack. For details, we again refer to the \textit{Materials and Methods} section and the original publication of the ARMAConv model~\cite{bianchi_graph_2021}.

The final model architecture is then set to seven ARMAConv layers with 512 channels per layer containing one GCS stack whose filtered response is computed over one iteration. 

\begin{figure}
\begin{adjustwidth}{-.35in}{-.35in}
\includegraphics[width=1.1\textwidth]{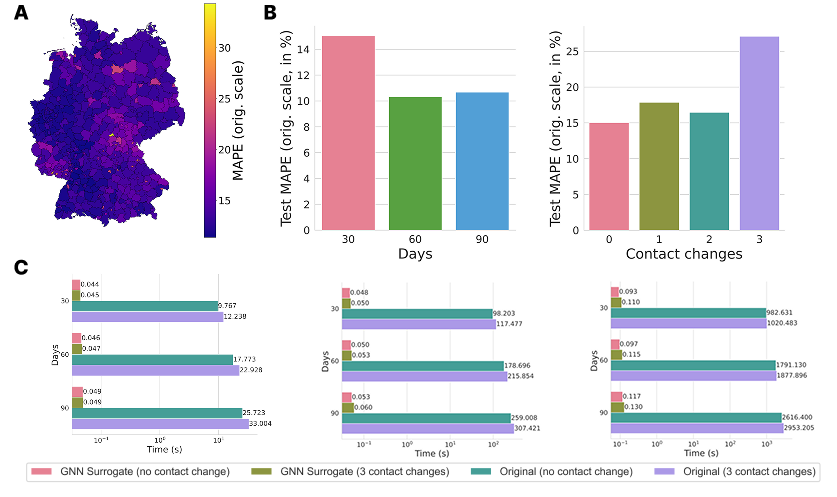}
\caption{\textbf{Test MAPE for final ARMAConv GNN and runtime comparison between GNN surrogate model and expert mechanistic model.} A) Test MAPE over the 400 counties. B) Test MAPE over prediction horizons of 30, 60, and 90 days without contact change and with 0, 1, 2, and 3 contact changes over 30 days horizon. C) Runtime comparison between final GNN and mechanistic expert model of~\cite{kuhn_assessment_2021} for one, ten, and 100 model executions with either zero or three contact changes over different prediction horizons.}
\label{fig:GNN_results}
\end{adjustwidth}
\end{figure}

Analogously to the simple neural networks employed for predictions without a spatial distribution, we tested the performance of the GNN on tasks with various prediction horizons (30, 60, 90 days) and numbers of contact change points (0, 1, 2, 3) within a time span of 30 days. 
Here, we used a classic 80-10-10 split of the data set into training in, validation, and test data.
The performance of the model was relatively constant for different prediction horizons. As depicted in~\cref{fig:GNN_results} (Panel A), we obtain test MAPEs between 10.33\,\% and 17.89\,\% for up to two contact changes on 30 days or without contact changes on 30, 60, and 90 days horizon. For a prediction horizon of 30 days and three contact changes, we obtain the highest MAPE with 27.11\,\%; see~\Cref{fig:GNN_results}.

\begin{figure}[h]
    \centering
    \includegraphics[width=0.7\linewidth]{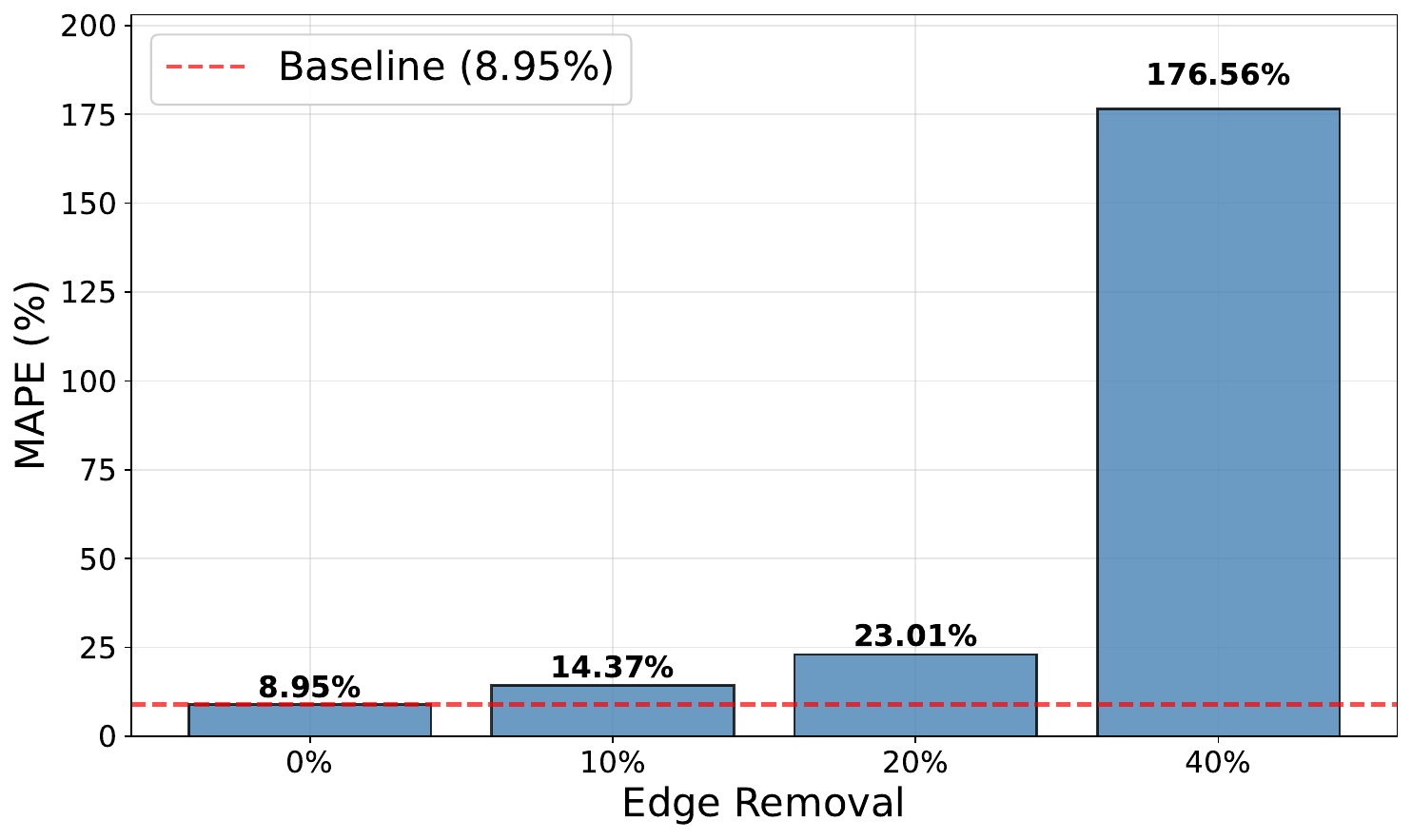}
    \caption{\textbf{Generalization of GNN surrogate under edge removal.} The bar chart shows MAPE (original scale) for four scenarios: baseline, 10\% edge removal, 20\% edge removal, and 40\% edge removal. Bars represent mean prediction error across 20 independent epidemic simulations per scenario using the pre-trained model without retraining. The red dashed horizontal line marks baseline performance on the original 400-county German mobility network.}
    \label{fig:gnn_generalization}
\end{figure}

\subsection{\MK{Generalization to modified networks}}

\MK{In this section, we briefly report on the generalization capabilities of the trained GNN. We use the model which was trained on a fixed graph topology where all $1,000$ training samples used the same mobility network.

We then created three modified mobility scenarios by randomly removing edges from the original network with \textbf{(b)} 10~\% edge removal, \textbf{(c)} 20~\% edge removal, and \textbf{(d)} 40~\% edge removal, always comparing against the baseline (0\% reduction). 

For each scenario, we generated 20 independent simulations using the mechanistic ODE metapopulation model with the modified mobility matrices using the persistent threat scenario. We then evaluated the pre-trained GNN model on these novel test sets and measured prediction accuracy using MAPE on the original scale. The results are presented in~\cref{fig:gnn_generalization} and demonstrate generalization capabilities for graph perturbations with 10 or 20~\% less edges and clear degradation as topological changes become substantial. With 10~\% edge removal, MAPE increases from 8.95\% to 14.37\%. The model maintains reasonable prediction accuracy, suggesting robustness to minor network disruptions (e.g., temporary mobility restrictions in specific regions). With 20~\% edge removal, MAPE rises to 23.01\%. While noticeable, predictions may still provide useful scenario guidance with the GNN showing sensitivity to moderate topological changes. With 40~\% edge removal, MAPE jumps to 176.56\%, indicating high prediction errors and when nearly half of the mobility network is removed.

In summary, the surrogate is reasonably robust to modest mobility changes (e.g., 10 to 20~\% edge reduction for simulating localized travel restrictions) but sensitive to major structural changes of the mobility network. On the other side, the mechanistic model itself is robust to any topology change, as it simulates dynamics from first principles and the surrogate trades this flexibility for computational speed.}

\subsection{Execution Time Performance}

While the time and energy needed for training a surrogate model should also be taken into account, the execution time performance is the critical factor if the model is to be used with a graphical user interface (GUI) that needs to respond immediately. The best GNN surrogate model from the previous section yielded test MAPEs between approximately 10\,\% and 27\,\%, when predicting \MK{age-structured and} spatially resolved disease dynamics over the 400 counties of Germany. In this section, we present result that indicate how the surrogate model outperforms the expert mechanistic model in terms of execution times. Timings were measured on a GPU-accelerated Intel Xeon Scalable Processor ``Skylake'' Silver node with four NVIDIA Tesla V100 SXM2 GPUs with 16 GB, utilizing a single GPU for the measurements.

To allow for just-in-time, web-based, exploration of different scenarios by manually adapting one parameter at a time, the model requires one single model prediction for every parameter change. When predefining multiple scenarios in a joint space over different parameters, for instance, the implementation of interventions to be evaluated on different days (e.g., today, in three days, and in one week) and with varying strictness (e.g., restricting gatherings to ten, 20, or 50 persons), a batch of multiple predictions are needed. With a factor space in just two or three dimensions and two or three discrete values, we then need between four and 27 model predictions. In~\Cref{fig:GNN_results}, we thus provide the measured run times for one, ten, and 100 executions of the expert and the surrogate model, respectively. Firstly, we observe that the expert model runtime scaled--as expected--linearly with the number of days while the GNN surrogate model provided results in (almost) constant time independent of the horizon. Furthermore, the execution time of the expert model also scaled linearly with the number of executions (without parallelization). Again, the scaling of the GNN surrogate model is clearly superior. While the timings are similar for one and ten executions, it only doubles to tripled from ten to 100 model executions. 
The prediction times of both models are almost independent of the number of contact change points. Overall, the neural network speeds up the prediction process by 222 to 673 times for a single model execution and by 2046 to 5123 times for ten model executions, going up to a maximal speed-up factor of 28\,670 for 100 executions and 90 days prediction horizon. With less than or approximately 0.1 seconds for the surrogate model to be executed, it can be employed on-the-fly for up to 100 model predictions.  Although the absolute execution speed on less powerful hardware setups will be slower, we expect that the efficiency of the GNN surrogate model still allows for rapid and interactive exploration scenarios, even in resource-poor environments.

\section{Discussion}
\label{sec:discussion}

Surrogate models are a cost-efficient technique to approximate the behavior of complex models. Usually, the more complex the reference model is, the higher the requirement of computational resources. The goal of the surrogate model is to approximate the original model's behavior, achieving accuracies close to the original model, at substantially reduced computational cost and time.
In time-critical epidemic or pandemic situations, the evaluation of a large set of potential reactions is recommended, and swift reaction is crucial~\cite{kuhn_regional_2022}. 
Unfortunately, supercomputing resources might not be available on model execution or predictions might be needed by stakeholders which do not have access to supercomputing facilities. To overcome these barriers and provide just-in-time answers to urgent questions, (pre)trained, e.g., overnight, surrogate models can become of great aid for decision makers and substantially increase the impact of mathematical-epidemiological modeling. \MK{With a regularly calibrated mechanistic model that integrates most recent viral parameter knowledge, mechanisms, and changed contact behavior, the surrogate model complements classical, purely data-driven ML models through the learning step of the simulator.}

In this article, we present a surrogate model based on neural networks that is capable of handling temporal and spatial resolution. While several recent surrogate models are based on the concept of physics-informed neural networks~\cite{calzolari_deep_2021, sun_surrogate_2020,shi_physics-informed_2022},
our surrogate model is entirely data-driven and uses a ``physics''-based simulator only to create a wide range of reliable training data. On the one hand, this means that physical constraints such as population conservation are not considered explicitly. On the other hand, our approach offers a substantial advantage, as particular modifications to the original model do not require changes in the surrogate model architecture; only new training data has to be generated and the model has to be retrained, potentially using the previous model as a starting point. Since (epidemiological) parameters usually do not change on a daily basis, retraining is only necessary if these parameters change substantially. In future work, we will investigate if our surrogate model could further benefit from integrating ``physical'' constraints such as population conservation into the loss function. 

Although we use age-resolved models and disease parameters, contact reductions in this study are applied homogeneously to all age groups. As NPIs, such as school closures or remote work policies, often target particular age groups, heterogeneous contact reductions should be studied in future research. While an implementation of heterogeneously changing contact patterns would be straightforward, we have not investigated how this would affect the performance of the surrogate model. Similarly, we have not considered spatially heterogeneous interventions. Both scenarios will be considered in future research. Moreover, when increasing the complexity of the contact restrictions in those scenarios, we expect that a larger data set and perhaps a larger network would be required to obtain a similar model performance.
Another limitation of the current study is that we have not yet considered vaccination, waning immunity, or reinfection. This means that the model in its current form has not been validated for realistic, later stage epidemic situations or more input settings.

%\MK{Moreover, all experiments in this work rely on synthetic data generated by the mechanistic model. While the mechanistic model itself has been validated against real-world COVID-19 data in previous studies~\cite{kuhn_assessment_2021,kuhn_regional_2022}, demonstrating accurate reproduction of infection dynamics in German counties, we did not perform a retrospective validation of the surrogate against observed pandemic trajectories in this study. A direct comparison with real data would require precise calibration of the mechanistic model to specific historical conditions and intervention timings, which introduces additional sources of uncertainty independent of the surrogate's architecture.}

In this study, we first compare four different types of GNN layers using the binary adjacency matrix modeling the mobility between different spatial regions as input. Therefore, although already providing convincing results, none of the considered approaches takes the highly varying values for traveling and commuting activities between the different spatial regions into account. We expect that incorporating this information as edge weights in the input graph could further improve the performance. Promising directions for incorporating the edge weights could be
the CrystalConv~\cite{xie_crystal_2018} or XENetConv~\cite{maguire_xenet_2021} layers. However, due to hardware limitations, specifically GPU memory on a single NVIDIA Quadro RTX 6000, we were unable to train models incorporating large graphs with 40,000 edges. Addressing this limitation will be a focus of future work.

This study \MK{mainly} sheds light on the performance of surrogate model using the same graph connectivity in training and evaluation processes. \MK{We only briefly explored} the generalization potential of the GNN model to other graphs than the one used for training, which is one of the strengths of GNNs. \MK{We considered the mobility network of 400 German counties and removed different quantities of edges, observing that a modest topological change with 20~\% removed edges still yields acceptable prediction results. In particular, research on generalization with several realistic mobility networks and} edge weights will be an avenue for future research.

With certain limitations and directions for future work, we present a surrogate model using methods of artificial intelligence that naturally integrate spatial information. The presented GNN performs well on the considered data sets for an initial or persistent disease outbreak and nonpharmaceutical interventions on all 400 counties of Germany. It is able to make predictions with low MAPEs over time horizons between 30 and 90 days, integrating contact pattern changes of the population on up to three different days during the considered period.

Our surrogate model delivers \MK{an age-structured and} spatially resolved disease dynamics for a prediction horizon of several months and speeds up the prediction process substantially. As these results can be computed within a fraction of a second, our model is capable of being integrated in web-based application.

\section*{Acknowledgements}
\noindent The authors thank Martin Siggel for the initial idea on developing AI surrogate models. AS thanks Simon Vollendorf for exchange on neural network models and technical considerations to run them. This work was supported by the Initiative and Networking Fund of the Helmholtz Association (grant agreement number KA1-Co-08, Project LOKI-Pandemics), by the German Federal Ministry for Digital and Transport under grant agreement FKZ19F2211A (Project PANDEMOS) and the Deutsche Forschungsgemeinschaft (DFG, German Research Foundation) (grant agreement 528702961).

\section*{Data availability:}
All data and code used to generate the results presented in this work are publicly available as part of the MEmilio framework~\cite{Bicker_MEmilio_2025}. The specific implementation used in this study, including all relevant scripts and pre-trained models, can be found under a persistent doi at zenodo~\cite{schmidt_2025_15055792}.

\section*{Author Contributions}
\noindent\textbf{Conceptualization:} AH, MK\\
\noindent\textbf{Data Curation:} AS, HZ\\
\noindent\textbf{Formal Analysis:} AS, HZ, AH, MK\\
\noindent\textbf{Funding Acquisition:} MK \\
\noindent\textbf{Investigation:} AS, HZ, AH, MK\\
\noindent\textbf{Methodology:} AS, HZ, AH, MK\\
\noindent\textbf{Project Administration:} MK\\
\noindent\textbf{Resources:} MK \\
\noindent\textbf{Software:} AS, HZ \\
\noindent\textbf{Supervision:} AH, MK \\
\noindent\textbf{Validation:} AS, HZ, AH, MK \\
\noindent\textbf{Visualization:} AS, HZ, MK \\
\noindent\textbf{Writing – Original Draft:} AS, HZ, AH, MK \\
\noindent\textbf{Writing – Review \& Editing:} AS, HZ, AH, MK

\section*{Funding}
This work was supported by the Initiative and Networking Fund of the Helmholtz Association (grant agreement number KA1-Co-08, Project LOKI-Pandemics), by the German Federal Ministry for Digital and Transport (grant agreement FKZ19F2211A, Project PANDEMOS), and by the Deutsche Forschungsgemeinschaft (DFG, German Research Foundation) (grant agreement 528702961). The funders had no role in study design, data collection and analysis, decision to publish, or
preparation of the manuscript.

\section*{Competing Interests}
All authors declare to not have any competing interests.

% \bibliographystyle{elsarticle-num}
\bibliographystyle{naturemag}

\appendix
\section{Supplementary Figures}
\begin{figure}[H]
\begin{adjustwidth}{-.35in}{-.35in}
    \includegraphics[width=1.\textwidth]{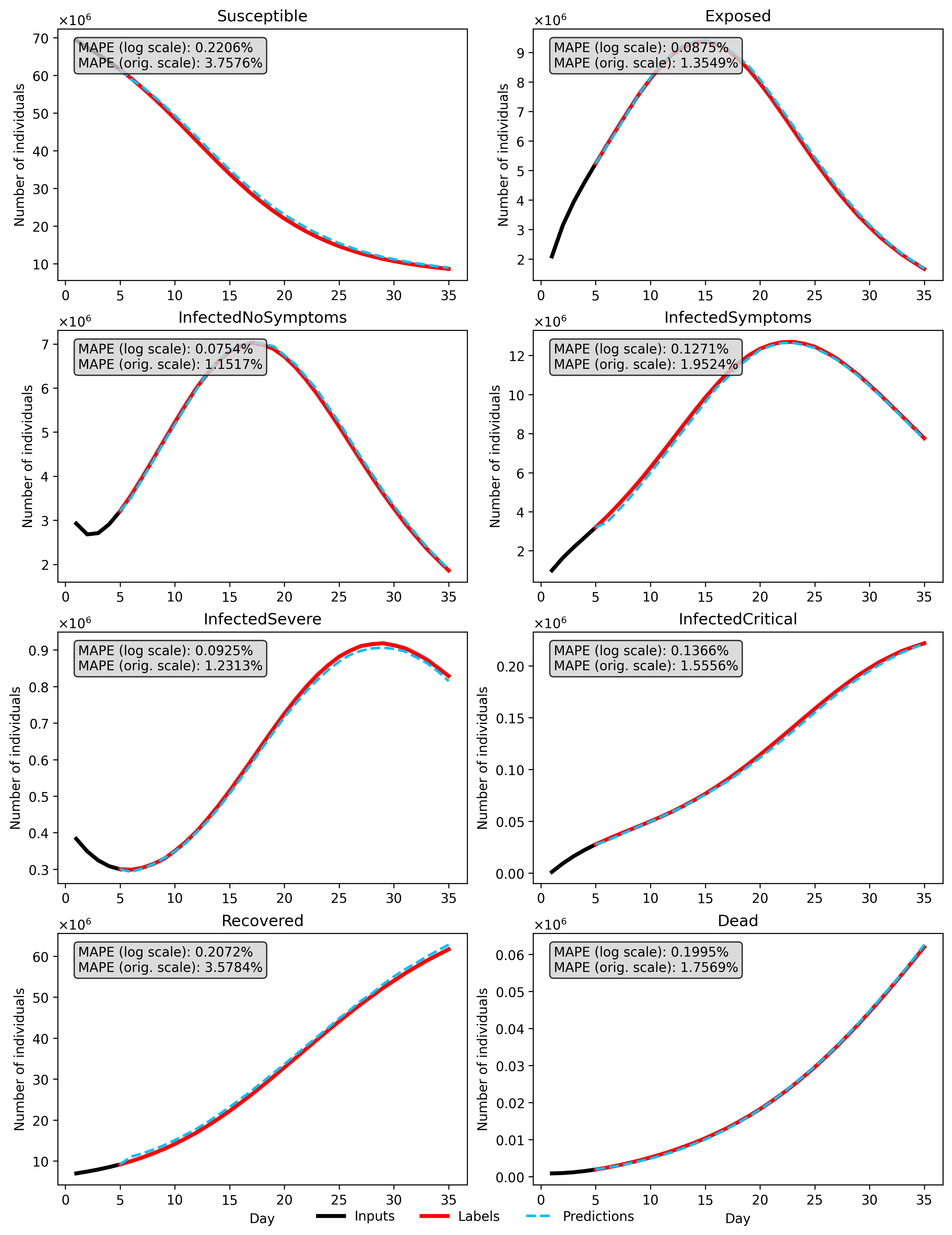}
	\caption{\textbf{LSTM predictions for the first trajectory of the model without stratification into age groups and corresponding MAPE values.} The particular prediction was chosen randomly and does neither represent a very good nor very bad prediction, it only allows to visually assess the trajectories against the corresponding MAPE values.}
	\label{fig:nns_trajectories_1}
\end{adjustwidth}
\end{figure}

\begin{figure} 
\begin{adjustwidth}{-.35in}{-.35in}
    \includegraphics[width=1.\textwidth]{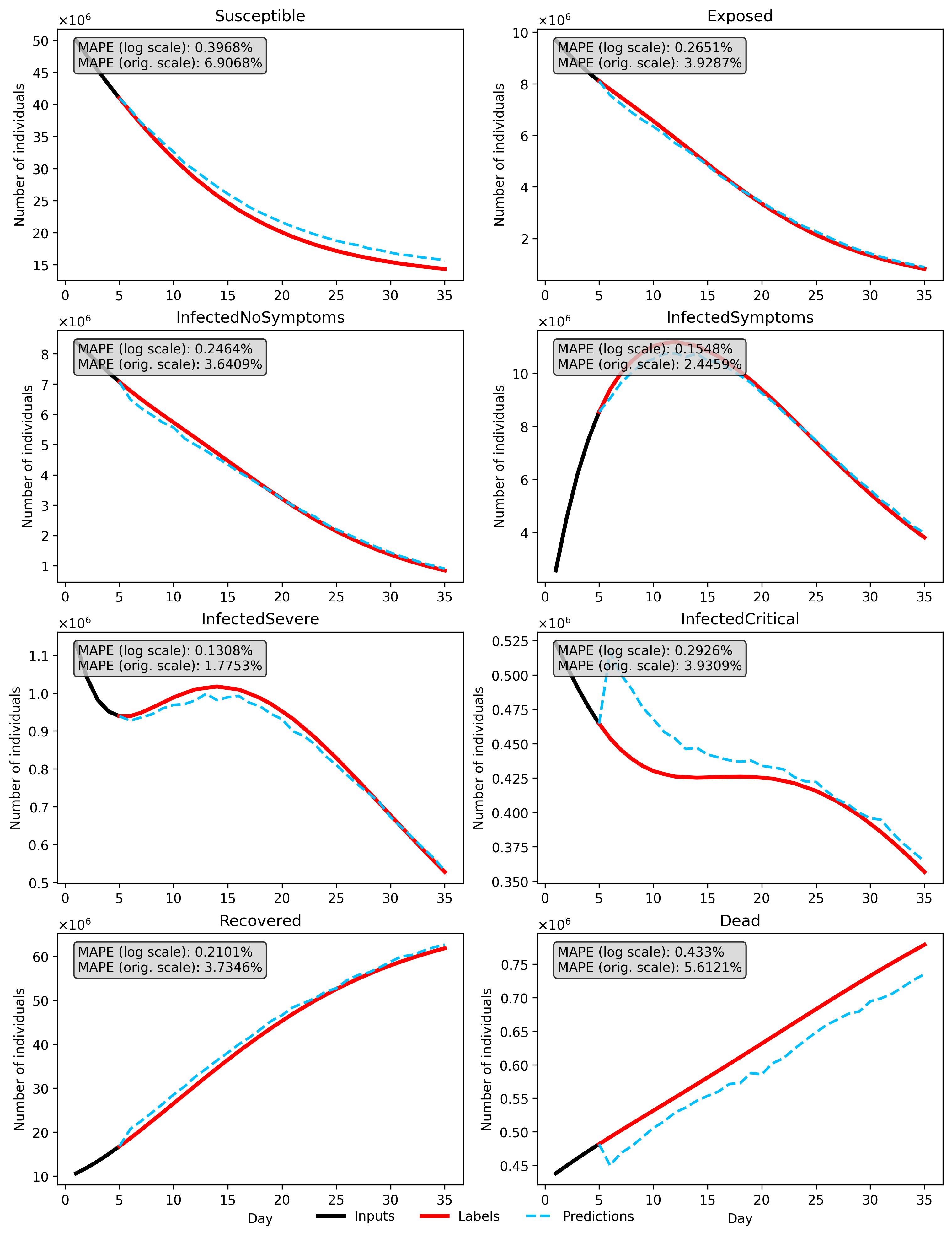}
	\caption{\textbf{LSTM predictions for the first trajectory of the model with stratification into age groups and corresponding MAPE values (summed over all age groups).} The particular prediction was chosen randomly and does neither represent a very good nor very bad prediction, it only allows to visually assess the trajectories against the corresponding MAPE values.}
	\label{fig:nns_trajectories_2}
\end{adjustwidth}
\end{figure}

\end{document}